\documentclass[11pt]{article}
\usepackage{paclic35}
\usepackage{times}
\usepackage{latexsym}
\usepackage{amsmath}
\usepackage{multirow}
\usepackage{url}
\usepackage{natbib}

\usepackage{microtype}
\usepackage{graphicx}
\usepackage{multirow}
\usepackage[usenames,dvipsnames]{xcolor}
\title{Exploring Conditional Text Generation\\for Aspect-Based Sentiment Analysis}

\author{Siva Uday Sampreeth Chebolu\textsuperscript{1}, Franck Dernoncourt\textsuperscript{2}, Nedim Lipka\textsuperscript{2}, Thamar Solorio\textsuperscript{1}  \\
        \textsuperscript{1}University of Houston, USA \\ \textsuperscript{2}Adobe Research  \\
        \textsuperscript{1}\texttt{\{sivauday.sampreeth8,thamar.solorio\}@gmail.com} \\ \textsuperscript{2}\texttt{\{franck.dernoncourt@gmail.com, lipka@adobe.com\}}
}

\begin{document}
\maketitle

\begin{abstract}
Aspect-based sentiment analysis (ABSA) is an NLP task that entails processing user generated reviews to determine (i) the target being evaluated, (ii) the aspect category to which it belongs, and (iii) the sentiment expressed towards the target and aspect pair.
In this article, we propose transforming ABSA into an abstract summary-like conditional text generation task that uses targets, aspects, and polarities to generate auxiliary statements.
To demonstrate the efficacy of our task formulation and a proposed system, we fine-tune a pre-trained model for conditional text generation task to get new state-of-the-art results on a few restaurant domain and urban neighborhoods domain benchmark datasets.
\end{abstract}

\section{Introduction}
\label{Sec: Introduction}
% Intro about ABSA and its recent progress

% What did TAS-BERT tackle?
%     They showed that joint modelling of aspect categories, targets and sentiments help each other in identifying the others and is necessary to obtain good results. They followed the BERT-pair-NLI-B task by Sun et al., to propose the joint modelling. Their main idea is based on the fact that the polarity depends both on aspect categories and targets in a given review sentence. 

% Limitation of TAS-BERT and BERT-pair-NLI-B?
%     Example: Always busy, but they are good at seating you promptly and have quick service. 
%     Labels: {service, SERVICE#GENERAL, positive}, {NULL, SERVICE#GENERAL, positive}
%     BERT-pair-NLI-B: {SERVICE#GENERAL, positive}, removed due to duplication
%     TAS-BERT: {service, SERVICE#GENERAL, positive}, removed due to conflicts between targets
    
%     As for the example, “service” and implicit target appear simultaneously.
%     When “service” and implicit target refer to the same aspect “SERVICE#GENERAL”, BERT-pair-NLI-B gives only one prediction, which is not completely correct either. Since a conflict between targets happens when the aspect and the sentiment are the same while one of multiple targets is NULL, our method fails to detect the opinion with implicit target. This failure is caused by data construction in the TAS-BERT method. 

Consumers and product makers/service providers both benefit from user-generated evaluations on e-commerce platforms.
Reading about previous customer experiences can assist future customers in making decisions. Reading about the aspects that create user feedback may help manufacturers and merchants develop ways to increase customer happiness.
Thousands of reviews covering various aspects and their corresponding opinions may be found in many situations.
Aspects can be a feature, a characteristic, or a behavior of a product or an entity, such as the ambiance of a restaurant, the performance of a laptop, the display of a phone, and so on. 
Aspect-based sentiment analysis is a task in which the sentiment for each aspect of an entity is determined.

This problem has two sub-problems: 1) aspect extraction (for example, sushi, pasta, and well-behaved staffs) and 2) finding the polarity toward each aspect.
Aspect extraction involves two sub-tasks: a) extracting aspect terms and b) categorizing/normalizing the extracted aspect terms into aspect categories.
Additionally, there are two sub-tasks in polarity detection: a) identify the polarity of an aspect word, b) determine the polarity of each category and build triplets from sentences [aspect term, aspect category, polarity].

To identify the aspect categories and their sentiments, \citet{14-Utilizing-Bert} recently turned this task into a sentence-pair classification task, such as Question-Answering or Natural Language Inference.
\citet{TargetAspectSentimentJD}, on the other hand, pointed out that sentiment is affected by both the aspect category and the aspect terms or targets in a review.
To overcome this challenge, they also proposed a joint model to extract the targets, aspect categories, and polarity by fine-tuning the BERT model with the auxiliary phrases (aspect-sentiment) generated by \citet{14-Utilizing-Bert}, as well as a sequence classification to identify the targets.

The drawback of such a formulation is that it is unable to detect implicit targets in situations where the same aspect-category-sentiment pair has explicit targets associated with it.
Take this review for example: \textit{"Always crowded, but they are good at seating you promptly and have quick service.}"
[\{service, SERVICE\#GENERAL, positive\}, \{NULL, SERVICE\#GENERAL, positive\}] are the actual labels for the targets, categories, and polarities.
\citet{14-Utilizing-Bert}, on the other hand, deleted the second opinion owing to redundancy, while \citet{TargetAspectSentimentJD} removed it due to target conflicts. 

Recently, Google released a unified framework (T5) \cite{T5} that transforms all text-based language problems into a text-to-text format, achieving state-of-the-art results on a variety of benchmarks including summarization, question answering, text classification, and more.
We suggest utilizing the multi-task method offered by \citet{TargetAspectSentimentJD} to reap the benefits of the T5 architecture. 
Deep generative models offer more expressive power, the ability to handle more forms of data, and the inherent benefit of being able to produce samples from the input text, all of which point to their suitability for the task at hand. 
The major benefit of conditional text generation is that it is mostly dependent on the content of an input text, which in our case contributes to the benefit of knowing which aspect category or polarity label to give and, finally, identify the target of those categories. 

In this work, we look at a few different ways to create an auxiliary phrase and turn ABSA into a conditional text generation task based on aspects and polarities, akin to abstractive summarization. 
On the ABSA challenge, we fine-tune the pre-trained models from T5, and BART (\cite{bart}) and reach new state-of-the-art results.
We also perform comparison tests to demonstrate that generation based on the multi-task framework is superior than specific task fine-tuned individually, implying that the improvement is due to both the pre-trained model and our approach. 
Furthermore, we investigate the performance of the T5-Encoder, which is pre-trained on a larger dataset than BERT, in the architecture proposed by \citet{TargetAspectSentimentJD}, to show that the improvement is due to the conversion to text generation rather than the larger dataset it is trained on. 
We observed that, this formulation enhanced implicit target identification while also removing the target conflict constraint in the sentence-pair classification approach proposed by \citet{TargetAspectSentimentJD}.

The following are our major contributions:
\begin{itemize}
    \item We construct the ABSA task as a conditional text generation problem, which comprises jointly extracting targets, categorizing the aspects into pre-defined categories, and their polarity.
    \item All of ABSA's sub-problems and associated tasks may be addressed in a sequence-to-sequence framework with our re-formulation, utilizing pre-trained models like T5.
    \item We do comprehensive tests on four public datasets, each of which comprises a subset of the ABSA subtasks, and show that our proposed framework outperforms current state-of-the-art techniques by a considerable margin.
\end{itemize}

%It might be good to be explicit about the contributions

% \section{Dataset and Metrics}
% \label{Sec: Dataset and Metrics}
% \input{Sections/2-Dataset-Metrics}

\section{Methodology}
\label{Sec: Methodology}

\subsection{Task Description}
    
    The TASD task aims to detect all triples (\textit{t, a, p}) that \textit{S} entails in the natural language meaning, where \textit{t} (called a target) is a subsequence of \textit{S}, \textit{a} is an aspect in \textit{A}, and \textit{p} is a sentiment polarity (simply called a sentiment) in \textit{P}, given a sentence \textit{S} consisting of \textit{n} words \textit{s1,..., sn}, a predefined set \textit{A} of aspects and a predefined set \textit{P} of polarities or sentiments. 
    %This sentence is hard to read. I would reverse the order and introduce P, S earlier
    The target \textit{t} can be \textit{NULL}, indicating that it is empty. An implicit target case is the name given to this situation. We name the triple (\textit{t, a, p}) an opinion since the overall objective of ABSA is to identify fine-grained sentiments. 
    Target-Detection (TD), Aspect-Detection (AD), Target-Sentiment Joint Detection (TSD), Aspect-Sentiment Joint Detection (ASD), and Target-Aspect Joint Detection (TAD) are the natural subtasks that arise from the TASD task. 
    From the example in Section \ref{Sec: Introduction}, there are 2 opinions (\{service, SERVICE\#GENERAL, positive\}) and (\{NULL, SERVICE\#GENERAL, positive\}). 
    With two positive sentiments, there is just one aspect for the two opinions, \textit{SERVICE\#GENERAL}, one explicit target, \textit{service}, and one implicit target, \textit{NULL}. 

    Apart from the tasks listed above, ABSA has two simpler tasks: one aims to classify sentiment according to a given aspect, as investigated in \citet{wang-etal-2016-attention,14-Utilizing-Bert}, 
    % xue-li-2018-aspect,
    and the other aims to classify sentiment according to a given target, as studied in \citet{zeng-ma-zhou-2019-target}.
    These tasks are not comparable to the TASD or its sub-tasks because they rely on prerequisite tasks like AD or TD to complete ABSA.

\subsection{Construction of auxiliary sentence}
    
        \begin{table*}[!hbt]
        \centering
        \resizebox{\textwidth}{!}{%
        \begin{tabular}{|l|l|l|l|}
        \hline
        \textbf{Task} & \textbf{Dataset} & \textbf{Pseudo Phrase} & \textbf{Pseudo Sentence}                                      \\ \hline
        \textbf{AD}  & All  & aspect                 & The review expressed opinion on {[}aspect{]}                  \\ \hline
        \textbf{ASD} & All & aspect$\sim$polarity   & The review expressed {[}polarity{]} opinion on {[}aspect{]}   \\ \hline
        \textbf{TD}  & SE-15, SE-16  & target                 & The review expressed opinion for {[}target{]}                 \\ \hline
        \textbf{TSD}  & SE-15, SE-16  & target$\sim$polarity   & The review expressed {[}polarity{]} opinion for {[}target{]}  \\ \hline
        \textbf{TAD}  & SE-15, SE-16  & target$\sim$aspect     & The review expressed opinion on {[}aspect{]} for {[}target{]} \\ \hline
        \textbf{TASD} & SE-15, SE-16 & target$\sim$aspect$\sim$polarity & The review expressed {[}polarity{]} opinion on {[}aspect{]} for {[}target{]} \\ \hline
        \end{tabular}%
        }
        \caption{Auxiliary Output Sentence formats and datasets on which the tasks are applied}
        \label{tab: auxiliary-sentence}
        \end{table*}
    
    To transform the ABSA task into a conditional text generation task, we explored the following two methods:

    \textbf{Pseudo output phrase}:  The sentence we obtain from a text generation model will only have the \textit{targets}, \textit{aspects}, and \textit{polarities} separated by a delimiter. 
    For the above example, we expect a model to generate (\textit{service $\sim$ SERVICE\#GENERAL $\sim$ positive $\sim$$\sim$ NULL $\sim$ SERVICE\#GENERAL $\sim$ positive}). 
    
    \textbf{Pseudo output sentence}: We will generate a complete sentence using a pseudo sentence format to obtain the labels for aspects, targets, and polarities. 
    The pseudo sentence for the same example will be \textit{"The review expressed [positive] opinion on [SERVICE\#GENERAL] for [service], [positive] opinion on [SERVICE\#GENERAL] for [NULL]"}. 
    
    The modifications in the format of the pseudo output phrase/sentence for all TASD subtasks are shown in Table \ref{tab: auxiliary-sentence}.
    Once we've created the auxiliary phrase/sentence for the output of a generation model, we can transform the ABSA task from a sentence classification task to a conditional text generation task.
    This is a necessary step that can significantly improve the ABSA task's experimental results.

\subsection{Fine-tuning pre-trained T5 model}
    
    T5 is a huge new neural network model that is trained on a combination of unlabeled text (the new C4 collection of English online text) and labeled data from popular natural language processing tasks, then fine-tuned separately for each task.
    It's an encoder-decoder model that translates all NLP problems to text-to-text. It requires an input sequence and a target sequence for training and is trained via teacher forcing.
    To do this, we feed the review text into the model as an input sequence and train it to generate a target sequence with one of the auxiliary sentence styles.
    To do this, we feed the review text into the model as an input sequence and train it to generate a target sequence with one of the auxiliary sentence styles.
    During the prediction phase, we simply provide the review text and evaluate the model output by extracting the targets, aspect categories, and polarity.

\subsection{T5-Joint and T5-Separate}
    
    \textbf{T5-Joint}: This is fine-tuning the T5 model for concurrently detecting targets, aspect categories, and polarities in order to complete all six tasks.
    As a fine-tuning target sequence, we utilize the format in the final row of Table \ref{tab: auxiliary-sentence}.
    
    \textbf{T5-Separate}: Here, we fine-tune the T5 model to solve the six tasks independently with a separate auxiliary sentence/phrase formats, as mentioned in Table \ref{tab: auxiliary-sentence}, for each task.

\section{Experiments}
\label{Sec: Expts}
\subsection{Datasets}
    The datasets that we utilized to assess our model are described in this section.    
    
    \textbf{SE-14}:
    SemEval-2014 Task4 dataset on restaurant reviews is used to evaluate the AD and ASD tasks. 
    Our model solves sub-task 3 (Aspect Category Detection) and the joint detection of aspect-sentiment tasks. 
    It is note that the sentiment score reported does not correspond to the subtask-4 (Aspect Category Sentiment Analysis) on the SE-14 dataset. This is because the formulation of the subtask-4 assumes that the gold aspect categories are already available for sentiment prediction, which is not the case for the task at hand. 
    % The categories are \textit{a} \in \textit{A} = \{food, service, price, ambience, anecdotes/miscellaneous\}, while the sentiment polarities are \textit{p} \in \textit{P} = \{positive, neutral, negative, conflict\} for SE-14 dataset. 
    % Most sentences in SE-14, on the other hand, have only one or multiple aspects with the same sentiment polarity, limiting ASD to sentence-level sentiment analysis.
    % MAMS-ACSA is a dataset published by \citet{MAMS-dataset} that has various aspect categories with distinct sentiment polarity in every sentence.
    
    \textbf{SE-15 and SE-16}:
    We also did tests on two other datasets in the restaurant domain, one of which is from SemEval-2015 (SE-15) Task 12, and the other is from SemEval-2016 (SE-16) Task 5. 
    On these two datasets, we assess all six tasks.
    We also assess TSD and TASD tasks for implicit targets, as described in \citet{TargetAspectSentimentJD}.
    % \citet{TargetAspectSentimentJD} reveals the limits of the majority of TASD subtasks. 
    % All implicit targets are ignored in both the TSD and TD tasks, which is critical because roughly a quarter of all opinions have implicit targets. 
    % Opinions grouped by the same sentence and the same target are referred to as target-sharing opinion groups, whereas opinions grouped by the same sentence and the same aspect are referred to as aspect-sharing opinion groups. 
    % Because the TSD task does not take aspects into account, no TSD method can accurately predict sentiments for multi-sentiment target-sharing opinion groups and vice-versa for the ASD task. 

    \textbf{SH}:
    We evaluate on the Sentihood (SH) dataset, which describes locations or neighborhoods of London and was collected from question answering platform of Yahoo. 
    The sentiment polarities are $p \in P$ = \{positive, negative and none\}, the targets are $t \in T$ = \{Location1, and Location2\}, and the aspect categories are $a \in A$ = \{general, price, transit-location, and safety\}. 
    The definition of target in this dataset, however, differs from that of the other benchmarks.
    The aspect-category is formed by combining the target and the aspect, resulting in eight aspect categories from the provided targets and aspects.
    As a result, we only evaluate the AD and ASD tasks on this dataset, rather than all six.

\subsection{Hyperparameters and Metrics}

    For all of our T5 model fine-tuning studies, we use the pre-trained T5-Base model.
    The encoder and decoder each include 12 transformer blocks, the size of the hidden layer is 768, and the pre-trained model has 220 million parameters.
    We utilize a maximum input sequence length of 128, a train batch size of 16, an evaluation batch size of 64, and a learning rate of 4e-5 for fine-tuning with a maximum of 50 epochs.
    Micro-F1 score and accuracy are used to evaluate our results for SE-14, SE-15, SE-16, and MAMS-ACSA, whereas macro-F1 score is used for the SH dataset.

\subsection{Comparison Methods}
    
    We compare our model with the following models:
    
    \textbf{SemEval-Top}
        The best scores in the SemEval contests are represented by SemEval-Top. These scores are available for three subtasks: AD, TD, and TAD.
        
    \textbf{MTNA}
        MTNA \cite{14-16-MTNA} is a multi-task model that identifies both aspect categories and targets using RNN and CNN. For both AD and TD, the research reported results on SE-14, SE-15, and SE-16.
        
    \textbf{Sentic LSTM + TA + SA}
        \citet{Sentic-LSTM} investigated Hierarchical attention, which first attends to the targets in a given review and then combines the aspect-embeddings and the output of a Sentic-LSTM network to apply another level of attention in order to produce target-aspect-specific sentence representations. They reported results on SE-15 and SH datasets on AD and ASD tasks. 
    
    \textbf{TAN}
        Instead of using aspect categories, \citet{14-16-TAN} created topic-specific sentence representations based on numerous topics. Furthermore, they employ a regularization term similar to Hu et al. to assure the uniqueness of the topics. They reported results on SE-14 and SE-16 for AD task only. 
    
    \textbf{BERT-pair-NLI-B}
        To fine-tune the pre-trained model using BERT for AD and ASD tasks, \citet{14-Utilizing-Bert} generate an auxiliary sentence from the aspect and transform ABSA to a sentence-pair classification task. They report results on SE-14 and SH datasets. 
        
    \textbf{baseline-1-f\_lex}
        \citet{baseline_f_lex} used a TASD pipeline technique. The study only presented findings on SE-15 for both TASD and its subtask ASD, and no source code is supplied. 
    
    \textbf{DE-CNN}
        \citet{DE-CNN} used a CNN model for TD. The paper only reported results on SE-16. 
        
    \textbf{THA + STN}
        A neural model for TD that includes a bi-linear attention layer and an FC layer \cite{THA-STN}. Both SE-15 and SE-16 were reported in the study.
    
    \textbf{BERT-PT}
        A BERT-based model to extract the targets (TD) \cite{BERT-PT}. Results are taken from \citet{TargetAspectSentimentJD} for SE-15 and SE-16. 
        
    \textbf{E2E-TBSA}
        Two layered recurrent neural networks are used in a unified framework to solve TSD in an end-to-end manner \cite{unified-TSD}: The upper one predicts the unified tags, while the bottom one predicts the auxiliary target boundaries. 
        Results are obtained from \citet{TargetAspectSentimentJD} for SE-15 and SE-16. 
        
    \textbf{DOER}
         A dual cross-shared RNN model for TSD \cite{DOER}. We used the results published in \citet{TargetAspectSentimentJD} for SE-15 and SE-16. 
    
    \textbf{TAS-BERT-SW-TO}
        \citet{TargetAspectSentimentJD} improved the BERT-pair-NLI-B architecture by detecting the presence of targets and extracting them using a TO tag-schema along with the aspect and sentiment of a review sentence. They reported results on SE-15 and SE-16 datasets for all the six tasks. 
        
    \textbf{TAS-BERT-SW-BIO-CRF}
        \citet{TargetAspectSentimentJD} improved the BERT-pair-NLI-B architecture by detecting the presence of targets and extracting them using a BIO tag-schema along with the aspect and sentiment of a review sentence. They reported results on SE-15 and SE-16 datasets for all the six tasks. 
    
    \textbf{QACG-BERT}
        \citet{QACG-BERT} create a CG- BERT that employs context-guided (CG) softmax- attention by first adapting a context-aware Transformer. Next, they present an enhanced Quasi-Attention CG-BERT model that learns a compositional attention model that enables subtractive attention.

    % \subsubsection{Exp-1: Aspect Detection (AD)}
    
    % \subsubsection{Exp-2: Aspect-Sentiment Detection (ASD)}

    % \subsubsection{Exp-3: Target Detection (TD)}

    % \subsubsection{Exp-4: Target-Sentiment Detection (TAD)}

    % \subsubsection{Exp-5: Target-Aspect Detection (TAD)}

    % \subsubsection{Exp-6: Target-Aspect-Sentiment Detection (TASD)}

\section{Results}
\label{Sec: Results}
\begin{table}[!htb]
    \centering
    \resizebox{\linewidth}{!}{%
        \begin{tabular}{|l|c|c|c|c|}
            \hline
            \multicolumn{1}{|c|}{\multirow{2}{*}{Model}} & \multirow{2}{*}{AD (Micro F1)} & \multicolumn{3}{c|}{ASD (Accuracy)}                   \\ \cline{3-5} 
            \multicolumn{1}{|c|}{} &       & 4-way & 3-way & 2-way \\ \hline
            SemEval-Top             & 88.58 & 82.90 & -     & -     \\
            MTNA                   & 88.91 & -     & -     & -     \\
            TAN                    & 90.61 & -     & -     & -     \\
            SCAN-BERT-AVE          & -     & 88.61 & -     & -     \\
            BERT-pair-NLI-B                              & 92.18               & 78.65 & 79.98 & 84.35 \\
            QACG-BERT              & 92.64 & 77.80  & 80.10  & 82.77 \\ \hline
            T5-Phrase-AD           & 85.30 & -     & -     & -     \\
            T5-Sentence-AD         & 92.50 & -     & -     & -     \\ \hline
            T5-Phrase-Joint-ASD    & 91.00 & 79.00 & 81.00 & 82.00 \\
            T5-Sentence-Joint-ASD                        & \textbf{93.34}      & \textbf{82.75}        & \textbf{84.50}        & \textbf{85.62 }       \\ \hline
        \end{tabular}%
    }
    \caption{Results for ASD task on SE-14 dataset. Micro F1-score is used for AD and Accuracy for ASD task.}
    \label{tab: Results-SE-14}
\end{table}

%
% Sentihood dataset results
% 
\begin{table}[!htb]
    \centering
    \resizebox{\linewidth}{!}{%
        \begin{tabular}{|l|c|c|}
            \hline
            \multicolumn{1}{|c|}{\textbf{Model}} & \multicolumn{1}{c|}{\textbf{AD (Macro F1)}} & \multicolumn{1}{c|}{\textbf{ASD (Accuracy)}} \\ \hline
            LSTM-Final            & 68.90          & 82.00                  \\
            Sentic LSTM + TA + SA & 78.20          & 89.30                  \\
            Dmu-Entnet            & 78.50          & 91.00                  \\
            BERT-pair-QA-M        & 86.40          & 93.60                  \\
            BERT-pair-QA-B        & 87.90          & 93.30                  \\
            QACG-BERT             & 89.70          & \textbf{93.80}         \\ \hline
            T5-Phrase-AD          & 89.68          & \multicolumn{1}{l|}{-} \\
            T5-Sentence-AD        & 88.75          & \multicolumn{1}{l|}{-} \\ \hline
            T5-Phrase-Joint-ASD   & 90.45          & 93.06                  \\
            T5-Sentence-Joint-ASD & \textbf{90.63} & 91.56                  \\ \hline
        \end{tabular}%
    }
    \caption{Results on SentiHood dataset.  Macro F1-score is used for AD and Accuracy for ASD task.}
    \label{tab: Results-SH}
\end{table}

% 
% SE-15 and SE-16 dataset results
% 
\begin{table*}[!htb]
    \centering
    \resizebox{\textwidth}{!}{%
        \begin{tabular}{lrrlrrlrr}
            \hline
            \multicolumn{3}{|c|}{\textbf{AD}} &
              \multicolumn{3}{c|}{\textbf{TD}} &
              \multicolumn{3}{c|}{\textbf{TAD}} \\
            \multicolumn{1}{|c}{\textbf{Method}} &
              \multicolumn{1}{c}{\textbf{SE-15}} &
              \multicolumn{1}{c|}{\textbf{SE-16}} &
              \multicolumn{1}{c}{\textbf{Method}} &
              \multicolumn{1}{c}{\textbf{SE-15}} &
              \multicolumn{1}{c|}{\textbf{SE-16}} &
              \multicolumn{1}{c}{\textbf{Method}} &
              \multicolumn{1}{c}{\textbf{SE-15}} &
              \multicolumn{1}{c|}{\textbf{SE-16}} \\ \hline
            \multicolumn{1}{|l}{SemEval-Top} &
              62.68 &
              \multicolumn{1}{r|}{73.03} &
              SemEval-Top &
              70.05 &
              \multicolumn{1}{r|}{72.34} &
              SemEval-Top &
              42.90 &
              \multicolumn{1}{r|}{52.61} \\
            \multicolumn{1}{|l}{MTNA} &
              65.97 &
              \multicolumn{1}{r|}{76.42} &
              MTNA &
              67.73 &
              \multicolumn{1}{r|}{72.95} &
              TAS-BERT-SW-BIO-CRF &
              63.37 &
              \multicolumn{1}{r|}{71.64} \\
            \multicolumn{1}{|l}{Sentic LSTM + TA + SA} &
              73.82 &
              \multicolumn{1}{r|}{-} &
              DE-CNN &
              - &
              \multicolumn{1}{r|}{74.37} &
              TAS-BERT-SW-TO &
              62.60 &
              \multicolumn{1}{r|}{69.98} \\
            \multicolumn{1}{|l}{TAN} &
              - &
              \multicolumn{1}{r|}{73.38} &
              THA+STN &
              71.46 &
              \multicolumn{1}{r|}{73.61} &
               TAS-BERT-large-SW-BIO-CRF&
              63.97 &
              \multicolumn{1}{r|}{72.45} \\
            \multicolumn{1}{|l}{BERT-pair-NLI-B} &
              70.78 &
              \multicolumn{1}{r|}{80.25} &
              BERT-PT &
              73.15 &
              \multicolumn{1}{r|}{77.97} &
               TAS-BERT-large-SW-TO&
              63.31 &
              \multicolumn{1}{r|}{70.94} \\
            \multicolumn{1}{|l}{TAS-BERT-SW-BIO-CRF} &
              76.34 &
              \multicolumn{1}{r|}{81.57} &
              TAS-BERT-SW-BIO-CRF &
              75.00 &
              \multicolumn{1}{r|}{81.37} &
               &
              \multicolumn{1}{l}{} &
              \multicolumn{1}{l|}{} \\
            \multicolumn{1}{|l}{TAS-BERT-SW-TO} &
              76.40 &
              \multicolumn{1}{r|}{82.77} &
              TAS-BERT-SW-TO &
              71.54 &
              \multicolumn{1}{r|}{78.10} &
               &
              \multicolumn{1}{l}{} &
              \multicolumn{1}{l|}{} \\
            \multicolumn{1}{|l}{TAS-BERT-large-SW-BIO-CRF} &
              77.18 &
              \multicolumn{1}{r|}{82.05} &
              TAS-BERT-large-SW-BIO-CRF &
              76.13 &
              \multicolumn{1}{r|}{81.99} &
               &
              \multicolumn{1}{l}{} &
              \multicolumn{1}{l|}{} \\
            \multicolumn{1}{|l}{TAS-BERT-large-SW-TO} &
              77.32 &
              \multicolumn{1}{r|}{83.64} &
              TAS-BERT-large-SW-TO &
              73.06 &
              \multicolumn{1}{r|}{79.22} &
               &
              \multicolumn{1}{l}{} &
              \multicolumn{1}{l|}{} \\
            
            \multicolumn{1}{|l}{BART-Phrase-AD} &
              71.13 &
              \multicolumn{1}{r|}{85.53} &
              BART-Phrase-TD &
              78.45 &
              \multicolumn{1}{r|}{79.51} &
               &
              \multicolumn{1}{l}{} &
              \multicolumn{1}{l|}{} \\
            
            \multicolumn{1}{|l}{T5-Phrase-AD} &
              72.96 &
              \multicolumn{1}{r|}{74.53} &
              T5-Phrase-TD &
              - &
              \multicolumn{1}{r|}{-} &
               &
              \multicolumn{1}{l}{} &
              \multicolumn{1}{l|}{} \\
              
            \multicolumn{1}{|l}{T5-Sentence-AD} &
              79.25 &
              \multicolumn{1}{r|}{83.75} &
              T5-Sentence-TD &
               78.46 &
              \multicolumn{1}{r|}{79.53} &
               &
              \multicolumn{1}{l}{} &
              \multicolumn{1}{l|}{} \\\hline
            
            \multicolumn{1}{|l}{BART-Phrase-ASD} &
              \textbf{80.70} &
              \multicolumn{1}{r|}{84.56} &
              BART-Phrase-TSD &
              76.29 &
              \multicolumn{1}{r|}{77.13} &
              BART-Phrase-TAD &
              63.42 &
              \multicolumn{1}{r|}{67.17} \\
            
            \multicolumn{1}{|l}{T5-Phrase-ASD} &
              \textbf{79.45} &
              \multicolumn{1}{r|}{84.05} &
              T5-Phrase-TSD &
              79.38 &
              \multicolumn{1}{r|}{83.32} &
              T5-Phrase-TAD &
              67.53 &
              \multicolumn{1}{r|}{74.07} \\
            \multicolumn{1}{|l}{T5-Sentence-ASD} &
              79.11 &
              \multicolumn{1}{r|}{\textbf{85.24}} &
              T5-Sentence-TSD &
              79.60 &
              \multicolumn{1}{r|}{82.02} &
              T5-Sentence-TAD &
              66.48 &
              \multicolumn{1}{r|}{73.02} \\
            \multicolumn{1}{|l}{BART-Phrase-TAD} &
              78.99 &
              \multicolumn{1}{r|}{84.28} &
              BART-Phrase-TAD &
              \textbf{80.73} &
              \multicolumn{1}{r|}{82.76} &
               &
              \multicolumn{1}{l}{} &
              \multicolumn{1}{l|}{} \\
            \multicolumn{1}{|l}{T5-Phrase-TAD} &
              77.43 &
              \multicolumn{1}{r|}{82.97} &
              T5-Phrase-TAD &
              \textbf{80.73} &
              \multicolumn{1}{r|}{82.76} &
               &
              \multicolumn{1}{l}{} &
              \multicolumn{1}{l|}{} \\
            \multicolumn{1}{|l}{T5-Sentence-TAD} &
              77.98 &
              \multicolumn{1}{r|}{84.62} &
              T5-Sentence-TAD &
              80.65 &
              \multicolumn{1}{r|}{\textbf{84.30}} &
               &
              \multicolumn{1}{l}{} &
              \multicolumn{1}{l|}{} \\ \hline
            \multicolumn{1}{|l}{BART-Phrase-Joint-TASD} &
              78.85 &
              \multicolumn{1}{r|}{84.19} &
              BART-Phrase-Joint-TASD &
              76.99 &
              \multicolumn{1}{r|}{76.06} &
              BART-Phrase-Joint-TASD &
              64.63 &
              \multicolumn{1}{r|}{66.49} \\
            \multicolumn{1}{|l}{T5-Phrase-Joint-TASD} &
              77.31 &
              \multicolumn{1}{r|}{83.10} &
              T5-Phrase-Joint-TASD &
              80.69 &
              \multicolumn{1}{r|}{83.51} &
              T5-Phrase-Joint-TASD &
              67.19 &
              \multicolumn{1}{r|}{\textbf{74.65}} \\
            \multicolumn{1}{|l}{T5-Sentence-Joint-TASD} &
              78.58 &
              \multicolumn{1}{r|}{82.97} &
              T5-Sentence-Joint-TASD &
              79.98 &
              \multicolumn{1}{r|}{83.54} &
              T5-Sentence-Joint-TASD &
              \textbf{67.72} &
              \multicolumn{1}{r|}{73.03} \\ \hline
             &
              \multicolumn{1}{l}{} &
              \multicolumn{1}{l}{} &
               &
              \multicolumn{1}{l}{} &
              \multicolumn{1}{l}{} &
               &
              \multicolumn{1}{l}{} &
              \multicolumn{1}{l}{} \\ \hline
            \multicolumn{3}{|c|}{\textbf{ASD}} &
              \multicolumn{3}{c|}{\textbf{TSD}} &
              \multicolumn{3}{c|}{\textbf{TASD}} \\
            \multicolumn{1}{|c}{\textbf{Method}} &
              \multicolumn{1}{c}{\textbf{SE-15}} &
              \multicolumn{1}{c|}{\textbf{SE-16}} &
              \multicolumn{1}{c}{\textbf{Method}} &
              \multicolumn{1}{c}{\textbf{SE-15}} &
              \multicolumn{1}{c|}{\textbf{SE-16}} &
              \multicolumn{1}{c}{\textbf{Method}} &
              \multicolumn{1}{c}{\textbf{SE-15}} &
              \multicolumn{1}{c|}{\textbf{SE-16}} \\ \hline
            \multicolumn{1}{|l}{baseline-1\_f\_lex} &
              - &
              \multicolumn{1}{r|}{63.50} &
              E2E-TBSA &
              53.00 (-) &
              \multicolumn{1}{r|}{63.10 (-)} &
              baseline-1\_f\_lex &
              - &
              \multicolumn{1}{r|}{38.10} \\
            \multicolumn{1}{|l}{BERT-pair-NLI-B} &
              63.67 &
              \multicolumn{1}{r|}{72.70} &
              DOER &
              56.33 (-) &
              \multicolumn{1}{r|}{65.91 (-)} &
              TAS-BERT-SW-BIO-CRF &
              57.51 &
              \multicolumn{1}{r|}{65.89} \\
            \multicolumn{1}{|l}{TAS-BERT-SW-BIO-CRF} &
              68.50 &
              \multicolumn{1}{r|}{74.12} &
              TAS-BERT-SW-BIO-CRF &
              66.11 (64.29) &
              \multicolumn{1}{r|}{75.68 (72.92)} &
              TAS-BERT-SW-TO &
              58.09 &
              \multicolumn{1}{r|}{65.44} \\
            \multicolumn{1}{|l}{TAS-BERT-SW-TO} &
              70.42 &
              \multicolumn{1}{r|}{76.33} &
              TAS-BERT-SW-TO &
              64.84 (65.02) &
              \multicolumn{1}{r|}{73.34 (71.02)} &
               TAS-BERT-large-SW-BIO-CRF &
               58.12 &
              \multicolumn{1}{r|}{67.19} \\
            \multicolumn{1}{|l}{TAS-BERT-large-SW-BIO-CRF} &
              69.12 &
              \multicolumn{1}{r|}{74.87} &
              TAS-BERT-large-SW-BIO-CRF &
              67.23 (66.09) &
              \multicolumn{1}{r|}{75.96 (76.42)} &
              TAS-BERT-large-SW-TO &
              58.53 &
              \multicolumn{1}{r|}{66.29} \\
            \multicolumn{1}{|l}{TAS-BERT-large-SW-TO} &
              68.95 &
              \multicolumn{1}{r|}{75.81} &
              TAS-BERT-large-SW-TO &
              63.99 (66.19) &
              \multicolumn{1}{r|}{72.74 (74.00)} &
              TAS-T5-SW-TO&
              52.08 &
              \multicolumn{1}{r|}{59.63} \\ \cline{1-6}
              
            \multicolumn{1}{|l}{BART-Phrase-ASD} &
              70.98 &
              \multicolumn{1}{r|}{73.80} &
              BART-Phrase-TSD &
              71.52 (70.80) &
              \multicolumn{1}{r|}{68.16 (68.34)} &
              TAS-T5-SW-BIO-CRF &
              53.29 &
              \multicolumn{1}{r|}{61.56} \\
            
            \multicolumn{1}{|l}{T5-Phrase-ASD} &
              \textbf{71.65} &
              \multicolumn{1}{r|}{78.09} &
              T5-Phrase-TSD &
              70.95 (71.46) &
              \multicolumn{1}{r|}{\textbf{78.03 (77.40)}} &
               &
              \multicolumn{1}{l}{} &
              \multicolumn{1}{l|}{} \\
            \multicolumn{1}{|l}{T5-Sentence-ASD} &
              71.00 &
              \multicolumn{1}{r|}{\textbf{79.00}} &
              T5-Sentence-TSD &
              71.04 (70.82) &
              \multicolumn{1}{r|}{76.75 (76.03)} &
               &
              \multicolumn{1}{l}{} &
              \multicolumn{1}{l|}{} \\ \hline
            \multicolumn{1}{|l}{BART-Phrase-Joint-TASD} &
              71.24 &
              \multicolumn{1}{r|}{75.71} &
              BART-Phrase-Joint-TASD &
              68.92 (68.91) &
              \multicolumn{1}{r|}{68.80 (68.00)} &
              BART-Phrase-Joint-TASD &
              58.53 &
              \multicolumn{1}{r|}{60.25} \\
            \multicolumn{1}{|l}{T5-Phrase-Joint-TASD} &
              70.13 &
              \multicolumn{1}{r|}{77.17} &
              T5-Phrase-Joint-TASD &
              \textbf{71.60 (72.16)} &
              \multicolumn{1}{r|}{77.07 (78.22)} &
              T5-Phrase-Joint-TASD &
              \textbf{61.42} &
              \multicolumn{1}{r|}{\textbf{69.85}} \\
            \multicolumn{1}{|l}{T5-Sentence-Joint-TASD} &
              70.55 &
              \multicolumn{1}{r|}{76.05} &
              T5-Sentence-Joint-TASD &
              71.57 (70.82) &
              \multicolumn{1}{r|}{76.50 (77.20)} &
              T5-Sentence-Joint-TASD &
              61.15 &
              \multicolumn{1}{r|}{67.48} \\ \hline
        \end{tabular}%
    }
    \caption{Results for SE-15 and SE-16  datasets for six tasks. }
    \label{tab: Res-15-16}
\end{table*}

Results on SE-14, SE-15 and SE-16, and SH datasets are presented in Table \ref{tab: Results-SE-14}, Table \ref{tab: Res-15-16}, and Table \ref{tab: Results-SH} respectively. 
Each experiment's T5-based scores are averaged over three runs. 
Each table's first section relates to the results of earlier studies on that dataset.
The ablation study for a subtask of the main task described in that table is the second portion. 
For instance, ASD is the core task in the SE-14 and SH datasets, whilst the AD is utilized as an ablation experiment to assess the relevance of joint detection. 
Similarly, the main task in the SE-15 and SE-16 datasets is TASD, with the subtasks ASD, TSD, and TAD being examined. 
We report AD and ASD, and TD and TSD, for ASD and TSD respectively. 
Further, we report AD and TD scores for TAD. 
Finally, we provide the joint detection scores of each job for the specified dataset in the table's last section. 
The remainder of this section will use a task-based classification to describe each of those results.

% \subsection{Aspect Detection}

% \subsection{Aspect Sentiment Detection}

% \subsection{Target Detection}

% \subsection{Target Sentiment Detection}

% \subsection{Target-Aspect Detection}

% \subsection{Target-Aspect Sentiment Detection}

On the Restaurant and Urban Neighborhood domain datasets, our conditional text generation-based model consistently outperforms all attention-based and transformer-based approaches. 
These findings show that our technique correctly interprets the input language in order to generate the appropriate aspect categories, targets, and polarities. 
Furthermore, the T5-based studies demonstrate that performance on the complex implicit opinions is much enhanced. 
Our ablation task experiment and primary task findings (TASD), which included AD and TD, beat the TAS-BERT model considerably.
Furthermore, we can see that joint modeling experiments produce better results than single-task studies.
Moreover, jobs that included target identification improved by at least 3\%. 

On both ablation and primary tasks, the pseudo-sentence based formulation beat the phrase-based formulation in the SE-14 dataset.
The trend is inconsistent on the SE-15 and SE-16 datasets, though. The phrase-based formulation outperformed the sentence-based formulation on the majority of target-detection tasks. The phrase-based generation, for example, had the best performance on the TD, TSD, and TASD tasks on both datasets.
On a high level, phrase-based formulation did better on the SE-15 dataset, whereas sentence-based formulation performed better on aspect-detection related tasks in SE-16.  
It is possible that the longer sequence that had to be generated for the sentence-based formulation with multiple opinions had an impact on performance on target-related tasks. 

    % \usepackage{multirow}
    % \usepackage{graphicx}

\begin{table*}[]
        \centering
        \resizebox{\textwidth}{!}{%
        \begin{tabular}{|l|l|l|l|l|l|}
            \hline
            \multicolumn{1}{|c|}{\multirow{2}{*}{\textbf{Id}}} &
            \multicolumn{1}{|c|}{\multirow{2}{*}{\textbf{Review Sentence}}} &
              \multicolumn{1}{c|}{\multirow{2}{*}{\textbf{Gold Annotations}}} &
              \multicolumn{3}{c|}{\textbf{Model \& Predictions}} \\ \cline{4-6} 
            \multicolumn{1}{|c|}{} &
              \multicolumn{1}{c|}{} &
              \multicolumn{1}{c|}{} &
              \multicolumn{1}{c|}{\textbf{TAS-BERT}} &
              \multicolumn{1}{c|}{\textbf{BART}} &
              \multicolumn{1}{c|}{\textbf{T5}} \\ \hline
            S1 &
              \begin{tabular}[l]{@{}p{5cm}@{}}
              My g/f and I both agreed the food was very mediocre especially considering the price. \end{tabular}&
              \begin{tabular}[l]{@{}p{5cm}@{}}
                \{FOOD\#PRICES, food, negative\} \\ \\ \{FOOD\#QUALITY, food, negative\} 
              \end{tabular}&
              \color{Maroon}{\begin{tabular}[l]{@{}p{5cm}@{}}
                \textbf{\{FOOD\#PRICES, NULL, negative\}} \\ \\ \textbf{\{FOOD\#QUALITY, food, neutral\}} 
              \end{tabular}} &
              \color{OliveGreen}{\begin{tabular}[l]{@{}p{5cm}@{}}
                \{FOOD\#PRICES, food, negative\} \\ \\ \{FOOD\#QUALITY, food, negative\} 
              \end{tabular}}&
              \color{OliveGreen}{\begin{tabular}[l]{@{}p{5cm}@{}}
                \{FOOD\#PRICES, food, negative\} \\ \\ \{FOOD\#QUALITY, food, negative\} 
              \end{tabular}} \\ \hline
            S2 &
              \begin{tabular}[l]{@{}p{5cm}@{}}
              Amazing Spanish Mackeral special appetizer and perfect box sushi (that eel with avodcao -- um um um). \end{tabular}&
              \begin{tabular}[l]{@{}p{5cm}@{}}\{FOOD\#QUALITY, Spanish Mackeral special appetizer, positive\}\\ \\ \{FOOD\#QUALITY, box sushi, positive\}\\ \\ \{FOOD\#QUALITY, eel with avodcao, positive\}\end{tabular} &
              \begin{tabular}[l]{@{}p{5cm}@{}}\color{OliveGreen}{\{FOOD\#QUALITY, Spanish Mackeral special appetizer, positive\}}\\ \\ \color{Maroon}{\textbf{\{FOOD\#QUALITY, perfect box sushi (that eel with avodcao, positive\}}} \\ \\ -- \end{tabular} &
              \begin{tabular}[l]{@{}p{5cm}@{}}\color{OliveGreen}{\{FOOD\#QUALITY, Spanish Mackeral special appetizer, positive\}}\\ \\ -- \\ \\ -- \end{tabular} &
              \color{OliveGreen}{\begin{tabular}[l]{@{}p{5cm}@{}}\{FOOD\#QUALITY, Spanish Mackeral special appetizer, positive\}\\ \\ \{FOOD\#QUALITY, box sushi, positive\}\\ \\ \{FOOD\#QUALITY, eel with avodcao, positive\}\end{tabular}} \\ \hline
            S3 &
              \begin{tabular}[l]{@{}p{5cm}@{}}
              Don't leave the restaurant without it. \end{tabular}&
              \begin{tabular}[l]{@{}p{5cm}@{}}\{FOOD\#QUALITY, NULL, positive\} \end{tabular}&
              \color{Maroon}{\begin{tabular}[l]{@{}p{5cm}@{}}\textbf{\{RESTAURANT\#GENERAL, NULL, positive\}} \end{tabular}}&
              \color{Maroon}{\begin{tabular}[l]{@{}p{5cm}@{}}\textbf{\{RESTAURANT\#GENERAL, NULL, negative\}} \end{tabular}}&
              \color{Maroon}{\begin{tabular}[l]{@{}p{5cm}@{}}\textbf{\{RESTAURANT\#GENERAL, NULL, positive\}} \end{tabular}}\\ \hline
            S4 &
              \begin{tabular}[l]{@{}p{5cm}@{}}
              It was absolutely amazing. \end{tabular}&
              \begin{tabular}[l]{@{}p{5cm}@{}}\{RESTAURANT\#GENERAL, NULL, positive\} \end{tabular}&
              \color{Maroon}{\begin{tabular}[l]{@{}p{5cm}@{}}\textbf{\{FOOD\#QUALITY, NULL, positive\}} \end{tabular}}&
              \color{Maroon}{\begin{tabular}[l]{@{}p{5cm}@{}}\textbf{\{FOOD\#QUALITY, NULL, positive\}} \end{tabular}}&
              \color{Maroon}{\begin{tabular}[l]{@{}p{5cm}@{}}\textbf{\{FOOD\#QUALITY, NULL, positive\}} \end{tabular}}\\ \hline
            S5 &
              \begin{tabular}[l]{@{}p{5cm}@{}}
              But the space is small and lovely, and the service is helpful \end{tabular} &
              \begin{tabular}[l]{@{}p{5cm}@{}}
                \{AMBIENCE\#GENARAL, space, positive'\}\\ \\ \{SERVICE\#GENERAL, service, positive'\}
              \end{tabular} &
              \begin{tabular}[l]{@{}p{5cm}@{}}
                \color{Maroon}{\textbf{\{AMBIENCE\#GENARAL, space, negative'\}}} \\ \\ \color{OliveGreen}{\{SERVICE\#GENERAL, service, positive'\}}
              \end{tabular} &
              \begin{tabular}[l]{@{}p{5cm}@{}}
                \color{Maroon}{\textbf{\{AMBIENCE\#GENARAL, space, negative'\}}} \\ \\ \color{OliveGreen}{\{SERVICE\#GENERAL, service, positive'\}}
              \end{tabular}  &
              \color{OliveGreen}{\begin{tabular}[l]{@{}p{5cm}@{}}
                \{AMBIENCE\#GENARAL, space, positive'\}\\ \\ \{SERVICE\#GENERAL, service, positive'\}
              \end{tabular}} \\ \hline
            
        \end{tabular}%
        }
        \caption{Qualitative Analysis. 
        {Wrong predictions are highlighted in \color{Maroon}{\textbf{Red}}.} 
        {Correct predictions are highlighted in \color{OliveGreen}{Green}}}
        \label{tab: qual-analysis}
    \end{table*}

\subsection{Qualitative Analysis}
    % We present a qualitative analysis of the results in Table \ref{tab: qual-analysis}. 
    % We considered different scenarios where several of the given models make wrong predictions on the input review sentence for the TASD task from SE-16 dataset test set. 
    % The ones represented by the green color are correct predictions, whereas the ones with red color are wrong predictions. 
    We observed different types of errors that occurred due to wrong predictions from the models. 
    We present a few case studies of these errors in Table \ref{tab: qual-analysis}, which we discuss in this section. 
    The first error made by the models is due to the insufficient context provided by the review sentence. For example, the sentence S3 in Table \ref{tab: qual-analysis} is only a part of an entire review that has 
    "\textit{Green Tea creme brulee is a must!}" as the previous sentence. 
    The predicted aspect category \textit{RESTAURANT\#GENERAL} is true when we consider the S3 independently. However, when we take the previous sentence into context, the gold annotation will be valid. 
    Similar is the case with S4, for which the models predict \textit{FOOD\#QUALITY} as an aspect category, assuming that the word \textit{it} refers to some food item. 
    This drawback historically exists in the NLP problems, and using coreference resolution is one solution to overcome this. 
    
    The second error is not correctly identifying the targets. In the sentence S1, the target \textit{food} belongs to both the opinions \textit{mediocre} and the \textit{price}. TAS-BERT model wasn't able to identify the target \textit{food}, which is farther away from the opinion on price, for the aspect category \textit{FOOD\#PRICES}. This can be attributed to the performance improvement in the target detection by the generation models. 
    However, in sentence S2, the BART-based model wasn't able to identify the second target completely. Even though TAS-BERT could identify a second target, it was not able to recognize that they are two separate targets and should have been two different opinions. The T5-based model can identify all the targets, aspects, and sentiments correctly. This is the third error type that we observed during our analysis. 
    
    The last error we noticed is that, even though the models identified the correct target-aspect pair, the sentiment given to that pair is wrong. For example, in sentence S5, all three methods identified the target-aspect pair correctly. However, the TAS-BERT and BART-based models made wrong sentiment predictions. We can also attribute these wrong sentiment predictions to the difference in the performance of the respective models on TAD, and TASD tasks from Table \ref{tab: Res-15-16}. 

\subsection{Comparison of BERT, BART, and T5}
    BART-based generation outperform BERT-based approaches on all tasks in the SE-15 dataset and several tasks in the SE-16 dataset, as shown in Table \ref{tab: Res-15-16}. 
    With the exception of aspect identification on the SE-15 dataset, T5-based generation consistently beat BART-based techniques in all situations.
    Furthermore, we do not report the results of the BART model's pseudo-sentence-based generation because it performed badly when compared to the phrase-based formulation due to it's weak generalization capacity to the format we used for the predictions.
    
    We wanted to make sure that the T5 model's improvements were attributable to our formulation and use of conditional text generation to address the problem, rather than the dataset's size or the number of steps it was pre-trained on.
    As a result, we conducted an experiment in which the T5-Encoder was used in place of the BERT-Encoder in the TAS-BERT model.
    For the SE-15 and SE-16 datasets, the results are only presented for the primary TASD job.
    It is clear that our model outperformed the TAS-T5-Encoder-based model substantially.
    This might be because T5 has been pre-trained on a text-to-text formulation rather than encoder-only classification tasks. 
    Furthermore, we used BERT-large, which has 340M parameters, to offer a fair comparison to the T5 model, which has 220M parameters vs 110M for BERT-base.
    Table \ref{tab: Res-15-16} shows that BERT-large, with more than twice the amount of parameters as BERT-base, did not produce compelling improvements.
    With more than 1.5 times the amount of parameters, T5 surpassed the BERT-large model once more.
    This demonstrates that the amount of parameters has no bearing on T5's performance.

\section{Related Work}
\label{Sec: Related Works}
In this section, we take a look at the background of conditional text generation based pre-trained models and existing work on the different sub-tasks of ABSA. 

\subsection{Conditional Text Generation}
    Text summarization \cite{abs-sen-summ}, reading comprehension and other applications of text generation from text to text may be found. Natural language text is generated for communicating, summarizing, or refining by comprehending the source text and getting its semantic representation. 
    In recent years, pre-trained language models have become a trend for various tasks in NLP, mainly for text generation, and conditional text generation. 
    From the available pre-trained models fro language generation \cite{roberta,gpt2,bart,T5}, we chose to use BART, and T5 in this work. 
    Both BART and T5 employ a conventional Transformer design (Encoder-Decoder), similar to the original Transformer model used for neural machine translation, but with some differences. 
    BART is more like a combination of BERT (which only uses the encoder) \cite{BERT} and GPT (which only uses the decoder) \cite{gpt2}. 
    The encoder utilizes a denoising goal similar to BERT, while the decoder tries to recreate the original sequence (autoencoder) token by token using the preceding (uncorrupted) tokens and the encoder output. 
    T5 uses common crawl web extracted text. 
    The pre-trained model is trained for $2^19$ steps using a BERT-base size encoder-decoder transformer with the denoising goal and the C4 dataset.
    T5 is pre-trained on following objectives: (1) language modeling (predicting the next word), (2) BERT-style objective (masking/replacing words with random words and predicting the original text), and (3) deshuffling ( which is shuffling the input randomly and try to predict the original text). 

\subsection{ABSA Sub-tasks}
    \subsubsection{Individual tasks}
        \textbf{AD}
        There has been a lot of work done on the AD task over the years, which seeks to detect aspects only. 
        SVM classifiers are trained to detect features in earlier research such as \citet{14-NRC-Canada,14-Vector-Space-SVM}. \citet{delayed-memory} provides neural models to help with the AD task performance. Different attention processes are incorporated to a neural model in the research \citet{14-16-MTNA,14-16-TAN,14-POS-GRU-Attention-CNN,15-CAN,15-context-aware-embeddings} to identify aspects more accurately. Recently, \citet{14-Utilizing-Bert,14-Enhancing-BERT,TargetAspectSentimentJD} used pre-trained models such as BERT to get encoded representations of different types of input to identify the aspect categories. 
        
        \textbf{TD}
        The TD task aims to extract targets only. Traditionally,   CRFs are used by \citet{jakob-gurevych-TD-extracting,yin-TD-Path-emb} to extract the targets. 
        \citet{liu-TD-syntactic} used syntactic patterns. 
        Lately, neural models such as CNN \citet{14-16-MTNA,GTRU} and RNN \cite{li-lam-2017-deep,14-16-MTNA,luo-2019a} are widely used in target extraction. 
        Different attention mechanisms are introduced by  \citet{wang-2017,THA-STN}, in addition to a neural model to extract targets. 
        \citet{TargetAspectSentimentJD} used a CRF and a softmax decoding strategy on the encoded token representation from the pre-trained model BERT to detect the targets.

    \subsubsection{Joint tasks}
        
        \textbf{ASD}
        The ASD task is intended to identify both aspects and sentiments at the same time. 
        \citet{schmitt-ASD} uses an end-to-end CNN model to handle this problem. This task is converted into a sentence-pair classification problem using citet14-Utilizing-Bert, which allows the BERT model to be fine-tuned.
        \citet{TargetAspectSentimentJD}, which was released recently, takes a similar technique to solving this task in a joint modeling scenario. 
        
        \textbf{TSD}
        The TSD job tries to identify targets and sentiments at the same time. \citet{mitchell-TSD-open} and \citet{zhang-TSD} simplify the TSD job to a sequence labeling problem, which is solved by a CRF decoder using hand-crafted linguistic features, respectively.
        Recently, neural models have been popular, and \citet{Li-unified-TSD} presents a unified model for the TSD problem that consists of two stacked LSTM networks. 
        \citet{DOER} proposes a dual cross-shared RNN model for TSD that incorporates sentiment lexicon and part-of-speech information. \citet{hu-TSD-open} offers a span-based pipeline architecture for solving the TSD challenge by fine-tuning a language model that has already been trained. 
        More recently, \citet{TargetAspectSentimentJD} solved this problem using a joint modelling objective by fine-tuning BERT on the input representation and using a softmax/CRF decoding mechanism on the encoded token representations. 
        
        \textbf{TAD}
        This task aims to detect the aspect categories and the target simultaneously. 
        There is very limited work on this task. In this task, detecting the aspect categories is treated as a classification task and the target detection as a sequence labelling task, which is acheived using the CRF decoding or softmax decoding on the token representations. \citet{14-16-MTNA} used a CNN and LSTM to jointly detect the targets and aspect categories in a sentence, while
        \citet{TargetAspectSentimentJD} used a joint modelling objective by fine-tuning a pre-trained language model BERT to obtain the token representation. 
        
        \textbf{TASD}
        TASD task aims to identify the triple: target, aspect categories, and polarities together. 
        This is also an under explored task with a few past works. \citet{baseline_f_lex} relied on available parsers and domain-specific semantic lexicons, but this method performs poorly as shown in our experiments. 
        Another method is the TAS-BERT joint modeling method that fine-tunes the pre-trained model BERT to solve the aspect-sentiment detection task using the classification token and the detecting the targets corresponding to the ASD using the token classification with CRF/softmax decoding. 
        
        All the above methods use a classification or sequence labeling method to solve the sub-tasks of ABSA. However, we employ a conditional text generation based method to solve the tasks simultaneously to capture both implicit and explicit targets corresponding to an aspect-sentiment pair effectively.

\section{Conclusion}
\label{Sec: Conclusion}
In this study, we use an unique generative framework to handle diverse ABSA challenges.
With a conditional text generation model, we tackle numerous ABSA sub-tasks, including many sentiment pair and triplet extraction tasks, by constructing the target sentences with our suggested pseudo-phrases and pseudo-sentences.
To demonstrate the efficacy of our proposed technique, we conducted comprehensive testing on several benchmark datasets from various domains spanning six ABSA sub-tasks. 

To demonstrate the efficacy of our formulation, we ran tests on a limited fraction of ABSA sub-tasks and variations.
As a result, this could be a first step toward investigating and converting most of ABSA's classification tasks, as well as other text classification tasks outside of ABSA, into a conditional text generative framework, in order to reap the benefits of this formulation and a better understanding of the given text.

\bibliographystyle{acl_natbib}
\bibliography{generation}

% \section{Supplementary Materials}
% \label{sec:supplementary}

%     \subsection{Appendices}
%     \label{subsec: appendix}
%         \input{Sections/7-Appendix}

\end{document}